\documentclass{article}
\usepackage[preprint]{neurips_2026}

\usepackage[utf8]{inputenc}
\usepackage[T1]{fontenc}
\usepackage{hyperref}
\usepackage{url}
\usepackage{booktabs}
\usepackage{amsfonts}
\usepackage{amssymb}
\usepackage{nicefrac}
\usepackage{microtype}
\usepackage{xcolor}
\usepackage{graphicx}
\usepackage{amsmath}
\usepackage{multirow}
\usepackage{makecell}
\usepackage[ruled,vlined]{algorithm2e}
\usepackage{colortbl}

\definecolor{gain}{rgb}{0.0,0.55,0.0}
\definecolor{loss}{rgb}{0.75,0.0,0.0}
\definecolor{rowg}{gray}{0.92}
\newcommand{\g}[1]{{\footnotesize\color{gain}$+#1$}}

\title{Recursive Self-Evolving Agents via Held-Out Selection}

\author{%
  Michael Nguyen \quad Quoc Nguyen \quad Paul Vuong \\
  School of Information Technology\\
  Monash University Malaysia\\
  \texttt{\{michael.nguyen, quoc.nguyen, paul.vuong\}@monash.edu}
}

\begin{document}
\maketitle

\begin{abstract}
LLM agents are increasingly improved \emph{without} weight updates, by evolving a
natural-language artifact -- reflections, workflows, playbooks, cheatsheets, or
optimized prompts -- that conditions a frozen policy. Such methods are typically
reported as wins on the single benchmark where they help. We study them
apples-to-apples and surface a sharper picture. We introduce \textbf{RSEA}
(Recursive Self-Evolving Agent), which carries a compact \emph{three-layer
natural-language state} -- an imperative \emph{strategy}, reusable \emph{skills},
and a procedural \emph{playbook} -- and across generations rewrites all three layers
from its own trajectories, committing a candidate only if it does not regress on a
disjoint held-out split (a \emph{strict keep-better} gate). Across four diverse
benchmarks (ALFWorld, GAIA, $\tau$-bench, WebShop) and six faithful baselines
(ReAct, Reflexion, GEPA, AWM, ACE, Dynamic Cheatsheet), all on one shared local
backbone, we find: (i)~\emph{no artifact universally wins} -- RSEA is the strongest
single-pass method on ALFWorld ($69.3\%$ vs.\ ReAct $64.6\%$, McNemar $p{=}0.015$;
$79.4\%$ with retry, best overall), while concrete-workflow induction (AWM) is best
on the strong-backbone tool-use tasks; (ii)~unguarded context evolution is
\emph{high-variance and unsafe} -- Dynamic Cheatsheet, which curates context online
with no held-out gate, is near-best on ALFWorld ($70.7\%$) yet \emph{collapses} on
WebShop (score $0.14$ vs.\ ReAct $0.43$); and (iii)~RSEA's strict held-out selection
is what makes recursive self-evolution \emph{monotone-safe}: it never significantly
underperforms the base agent on any benchmark, falling back to vanilla ReAct when
evolved context would hurt. Controlled ablations show that every layer of
the evolved state helps and that removing held-out selection causes severe
overfitting (a perfect in-sample score but a $33$-point drop to test). We argue the
reliability of context evolution comes less from the
artifact and more from the \emph{selection gate}, and release the full harness, the
six baseline re-implementations, and pre-registered manifests.
\end{abstract}

\section{Introduction}
A central paradigm for building self-improving LLM agents adapts the \emph{context}
rather than the weights: the policy is frozen and an evolving natural-language
artifact -- accumulated reflections, induced workflows, curated playbooks, online
cheatsheets, or optimized system prompts -- is injected at inference. Context
adaptation is attractive for concrete reasons: it is interpretable and auditable,
integrates new knowledge at runtime, transfers across model versions, and -- with
long-context inference and KV-cache reuse -- is increasingly cheap to serve.

Despite rapid progress, these methods are almost always evaluated \emph{in
isolation}: a single benchmark, weak or non-matched baselines, and a backbone or
decoding budget that varies between methods. This makes their \emph{relative} merits
and, crucially, their \emph{failure modes} hard to read. When we re-run six
representative methods on one shared backbone across four diverse agent benchmarks,
two uncomfortable facts emerge. First, \emph{which} artifact helps is
benchmark-dependent: a method that tops one benchmark is middling or harmful on
another. Second, and more importantly, evolving context is \emph{not} a free lunch.
We name the failure mode \emph{context distraction}: an injected artifact that
degrades an otherwise-capable base policy. Context distraction is not hypothetical
-- in our study an unguarded online method (Dynamic Cheatsheet) goes from
near-best on ALFWorld ($70.7\%$) to a WebShop score of $0.136$ versus the ReAct
baseline's $0.429$, a catastrophic regression caused entirely by the context it
added.

\textbf{We argue that the reliability of context evolution comes less from the
artifact being evolved and more from the \emph{selection gate} that decides whether
to commit it.} An evolution loop that commits whatever it produces inherits the
variance of its own rewrites; one that commits a change only when it improves
\emph{held-out} performance cannot do worse than not evolving at all. We make this
gate the design center of our method rather than an afterthought.

Concretely, we present \textbf{RSEA} (Recursive Self-Evolving Agent). RSEA carries a
compact three-layer natural-language state -- \emph{strategy} (an imperative
preamble), \emph{skills} (reusable sub-routines), and \emph{playbook} (procedures
distilled from successes) -- and across generations (i)~rolls the current state out
on a small evolve pool, (ii)~rewrites \emph{all three layers} from the resulting
trajectories, and (iii)~commits the candidate only if it does not regress on a
disjoint validation split, updating the frozen best state only on a \emph{strict}
improvement. Only the frozen best state is run on a held-out test split, so there is
no leakage. RSEA$_\text{R}$ injects this frozen prior into a per-task retry loop.

\textbf{Contributions.} (1)~A formulation of cross-task NL self-evolution centered on
a \emph{strict held-out keep-better} gate that makes the operation
\emph{monotone-safe} (\S\ref{sec:method}). (2)~A faithful, single-backbone
re-implementation of six classic and recent baselines on a shared harness
(\S\ref{sec:setup}). (3)~A four-benchmark study with multi-seed statistics and paired
tests, yielding an honest scope condition rather than a universal-win claim
(\S\ref{sec:alfworld}--\ref{sec:transfer}). (4)~Controlled ablations that isolate the
contribution of each state layer and quantify the cost of removing held-out
selection (\S\ref{sec:ablation}).

\textbf{Key findings.}
(a)~On ALFWorld (134 tasks $\times$ 5 seeds, 7B) the single-pass evolved prior
significantly beats ReAct, GEPA, and AWM, and RSEA$_\text{R}$ is best overall
($79.4\%$).
(b)~On the strong-backbone tool-use benchmarks no NL artifact dominates; RSEA is
statistically tied with ReAct (and never significantly worse), while AWM's concrete
workflows give the only consistent small lift.
(c)~Methods without a held-out gate are high-variance and can catastrophically
regress; RSEA's strict gate converts ``context evolution'' from a coin-flip into a
monotone-safe operation.
(d)~Ablations show every layer of the evolved state helps (the layers overlap rather
than being strictly complementary) and that removing held-out selection overfits the
evolve pool ($100\%$ in-sample, a $33$-point drop to test).

\begin{figure}[t]
  \centering
  \includegraphics[width=\linewidth]{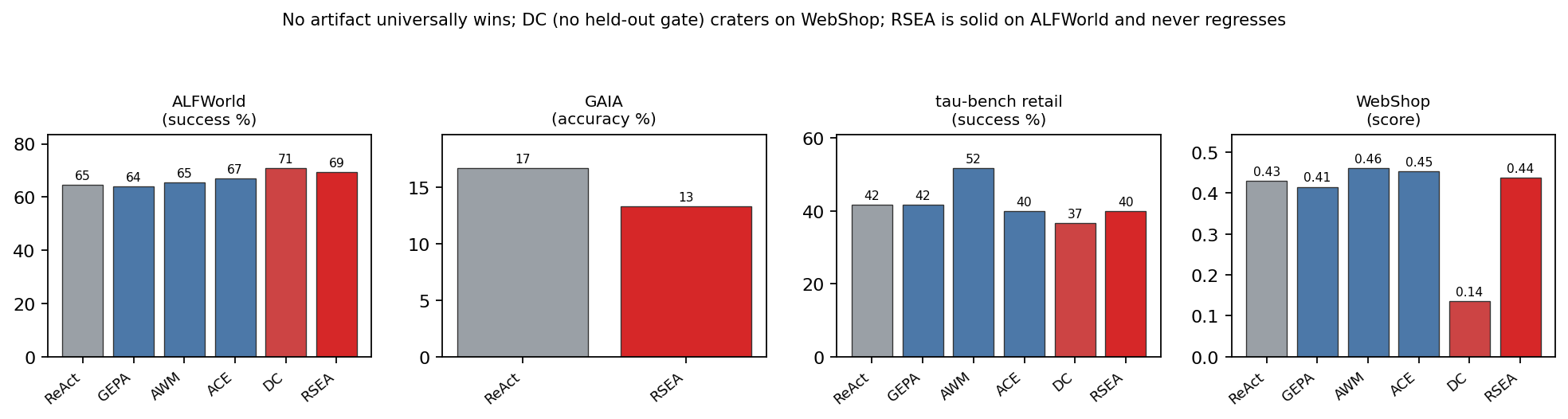}
  \caption{\textbf{No context-evolution artifact universally wins, and unguarded
  evolution is unsafe.} Single-pass methods across four benchmarks on one shared
  backbone (ALFWorld 7B; GAIA/$\tau$-bench/WebShop 30B). RSEA (red) is the strongest
  single-pass method on ALFWorld and never significantly underperforms ReAct (grey)
  elsewhere; AWM is best on the tool-use tasks; and Dynamic Cheatsheet -- which
  curates context online with \emph{no} held-out gate -- is near-best on ALFWorld yet
  \emph{collapses} on WebShop (0.14 vs.\ ReAct 0.43). RSEA's strict held-out gate is
  what makes evolution monotone-safe.}
  \label{fig:money}
\end{figure}

\section{Related Work}
\label{sec:related}
\paragraph{LLM agents that reason and act.} A now-standard recipe interleaves
free-form reasoning with tool/environment actions: ReAct~\citep{yao2023react} couples
chain-of-thought with acting, elaborated with deliberate search~\citep{yao2023tot},
self-critique and revision~\citep{madaan2023selfrefine}, learned tool
use~\citep{schick2023toolformer}, and grounded planning in embodied or open-ended
worlds~\citep{huang2022language, wang2023voyager}. Recent work strengthens the
reasoning substrate -- multi-perspective, semiotically grounded logical
reasoning~\citep{zhang2025ambiguity}, ``system-2'' vision-language-action
policies~\citep{song2025hume, fang2025dualvla}, and spatio-temporal chains of thought
for control~\citep{zeng2025FSDrive} -- while cognitive-architecture and survey
treatments organize the pieces into memory, planning, and action
modules~\citep{sumers2024coala, wang2024survey, park2023generative, yang2026toward}.
RSEA keeps this loop \emph{unchanged} and intervenes only through an evolved
natural-language state, isolating the contribution of context from scaffold.

\paragraph{Tool-use agents and tool learning.} A large body studies how agents select
and compose tools~\citep{qu2025tool}: mastering tools via self-driven
interaction~\citep{quexploration}, fine-grained supervision for tool-integrated
reasoning~\citep{qu2026matchtir}, similarity/dependency-aware experience networks for
multi-tool selection~\citep{zhang-etal-2025-toolexpnet}, self-critique under
tool-calling errors~\citep{huang2025critictool}, tree-search tool
planning~\citep{yang2026tooltree}, GUI grounding~\citep{lian2025ui}, and agentic
text-to-SQL~\citep{li2026sqlastra, su2026agenticsql}. Most closely, \emph{test-time
tool evolution} adapts an agent's tools at inference~\citep{lu2026beyond}. RSEA is
complementary: rather than evolving the tool set, it evolves a natural-language
\emph{strategy} over a fixed action/tool interface, and our $\tau$-bench/WebShop
results explain when such NL strategy adds little beyond an already-detailed tool API.

\paragraph{Prompt and context optimization.} A second line optimizes the prompt
itself: discrete instruction search~\citep{zhou2023ape}, LLM-as-optimizer
loops~\citep{yang2024opro}, declarative pipelines with compiled prompt/demonstration
optimization~\citep{khattab2024dspy, opsahlong2024mipro}, evolutionary
search~\citep{fernando2023promptbreeder}, and reflective genetic-Pareto
evolution~\citep{agrawal2025gepa}; prompts are also notoriously brittle to surface
form~\citep{cai2025does}. These methods optimize a \emph{single} flat artifact and
usually treat selection as an implementation detail. We instead hold the artifact
family fixed and show that the \emph{strictness} of held-out selection -- not the
sophistication of the optimizer -- governs whether evolution helps or hurts.

\paragraph{Self-evolving and self-improving agents.} Closest in spirit are agents
that improve from their own experience. Beyond context, this includes bootstrapping
reasoning from self-generated data~\citep{zelikman2022star} and searching over agent
code/topology~\citep{hu2024adas}. Within the agent loop, recent methods self-evolve
tool-use policies via blame-aware mutation and diversity-aware
selection~\citep{yang2026evotool}, evolve intrinsic skills in hierarchical
RL~\citep{li2026arise}, self-improve unified multimodal models from self-generated
supervision~\citep{han2026unicorn}, and evolve task-specific
prototypes~\citep{zhu2025pathology}. RSEA shares the self-evolution goal but is
deliberately weight-frozen and NL-state-based, and -- unlike most of these -- makes a
\emph{strict held-out selection gate} the center, which our ablations and transfer
results identify as the decisive ingredient for safety.

\paragraph{Reinforcement learning for LLM agents.} A parallel thread trains
agents/reasoners with RL: verifiable meta-reasoning rewards for long-horizon
agents~\citep{zhang2026rlvmr}, reasoning re-ranking agents~\citep{zhang2025rearank},
and RL for contextual integrity~\citep{lan2025contextual}. A recurring concern is the
design of the optimization signal and \emph{diversity collapse} under verifiable
rewards~\citep{li2025choice}, addressed via divergence choice~\citep{li2025choice},
reward-confidence correction~\citep{li2026right}, preference optimization with
priors~\citep{lan2025mappo}, and intrinsic self-reflection~\citep{li2025inspo}. Our
weight-frozen method avoids RL's cost and instability while still exploiting execution
feedback through the rewrite/selection loop.

\paragraph{Agent memory, experiential learning, and overfitting.} Closest to RSEA's
artifact are methods accumulating reusable NL experience: ExpeL~\citep{zhao2024expel},
Agent Workflow Memory~\citep{wang2024awm}, Agentic Context
Engineering~\citep{ace2025}, Dynamic Cheatsheet~\citep{suzgun2025dc}, and
memory-augmented agents with explicit stores~\citep{packer2023memgpt,
zeng2025janusvln}. Two failure modes from adjacent fields motivate our selection-first
view: \emph{catastrophic forgetting / collapse}, studied in continual
learning~\citep{chen-zeng-2025-prototype} and observed as logical-reasoning
collapse~\citep{zhang2026logical}; and classical \emph{overfitting}, controlled by
difficulty-aware reweighting and adversarial
robustness~\citep{zhou2022understanding, zhou2024boosting, zhou2024adversarial}.
RSEA's strict held-out gate is precisely a mechanism that prevents the evolved context
from overfitting or collapsing, which we demonstrate empirically.

\paragraph{Benchmarks, multi-agent systems, and the broader landscape.} We evaluate on
a deliberately diverse slate -- embodied text households~\citep{shridhar2021alfworld},
grounded web shopping~\citep{yao2022webshop}, open-ended assistants~\citep{mialon2023gaia},
tool-agent-user interaction~\citep{yao2024taubench}, interactive
coding~\citep{trivedi2024appworld}, and multimodal deep research~\citep{zeng2026vision}
-- a spread that surfaces the benchmark-dependence single-benchmark studies miss.
Beyond single agents, multi-agent and social-simulation systems coordinate many LLM
agents~\citep{zhang2025ga, zhang2026coupling}. Finally, our techniques sit within a
broad ecosystem of LLM and multimodal systems where self-improvement, evolution, and
robust reasoning recur: medical and scientific vision-language
models~\citep{zhu2025medeyes, wu2025bridging}, multimodal misinformation
detection~\citep{li2026s, li2025cmie}, vision-language alignment and
compositionality~\citep{Zhang_2024_CVPR, zhang2025assessing}, retrieval and query
expansion~\citep{zhang2024exploring, xie2025chat, xie2026hvd, xie2026conquer,
xie2026delving}, controllable generation and editing~\citep{xu2024headrouter},
automated optimization modeling~\citep{liu2026automated}, and efficient models via
knowledge distillation and compression~\citep{Lan_2025_ICCV, lan2026reco, Lan_2026_WACV}.

\section{Background and Motivation}
\label{sec:background}
\paragraph{Context adaptation.} Given a frozen policy $\pi_\theta$ and a benchmark of
tasks, context adaptation seeks an artifact $c$ (a string injected into the prompt)
that improves $\pi_\theta(\cdot\mid c)$ without touching $\theta$. Methods differ in
(i)~the \emph{form} of $c$, (ii)~the \emph{update operator} that edits $c$ from
execution feedback, and (iii)~the \emph{selection rule} that decides which $c$ to
keep. Table~\ref{tab:family} casts the methods we study in this $(\text{form},
\text{update}, \text{selection})$ view; the third column is where they differ most
and, we argue, where reliability is won or lost.

\paragraph{Limitations of existing methods.} Reflexion~\citep{shinn2023reflexion}
adapts \emph{within} a single task across retries but carries nothing across tasks.
Prompt/context optimizers carry a cross-task artifact but optimize a \emph{single}
form: GEPA~\citep{agrawal2025gepa} a flat prompt via reflective Pareto evolution; AWM
\citep{wang2024awm} a list of induced workflows; ACE~\citep{ace2025} an itemized
bullet playbook with add-only deltas; Dynamic Cheatsheet~\citep{suzgun2025dc} a
single cheatsheet curated \emph{online}. Two issues recur. \textbf{Single-form
artifacts} conflate complementary kinds of knowledge -- a high-level strategy, a
reusable sub-routine, and a concrete procedure are not interchangeable, yet most
methods fold them into one string. \textbf{Unguarded commitment} is the more
damaging: methods that append or rewrite without a held-out check (notably online
curation) commit context that can \emph{distract} the base policy, and -- because the
same mechanism that helps when context is useful is unconstrained when it is not --
they are high-variance across benchmarks (\S\ref{sec:transfer}). RSEA targets both:
a three-layer form, and a strict held-out gate.

\begin{table}[t]
  \centering
  \caption{Context-adaptation methods as (form, update operator, selection rule).
  The selection rule -- whether a candidate is committed only after improving
  \emph{held-out} performance -- is what we argue governs reliability.}
  \label{tab:family}
  \footnotesize
  \begin{tabular}{llll}
    \toprule
    \textbf{Method} & \textbf{Artifact form} & \textbf{Update operator} & \textbf{Selection rule} \\
    \midrule
    Reflexion & per-task reflection & verbal self-feedback & none (within-task retry) \\
    GEPA & flat prompt & reflective mutation & Pareto on val minibatch \\
    AWM & workflow list & induction from successes & none (induce \& inject) \\
    ACE & bullet playbook & add-only deltas & val keep-best (non-strict) \\
    Dynamic Cheatsheet & single cheatsheet & online rewrite & \textbf{none (no held-out gate)} \\
    \textbf{RSEA (ours)} & \textbf{3-layer state} & holistic rewrite of all layers & \textbf{strict held-out keep-better} \\
    \bottomrule
  \end{tabular}
\end{table}

\section{Method: RSEA}
\label{sec:method}
\begin{figure}[t]
  \centering
  \includegraphics[width=0.94\linewidth]{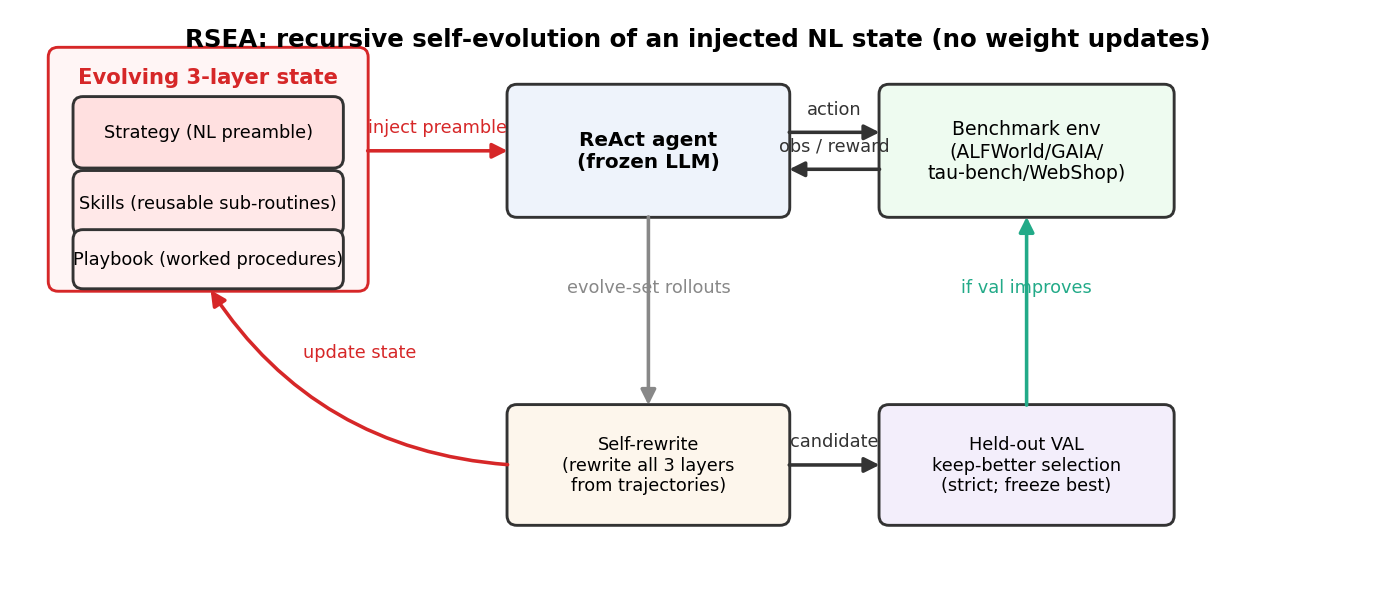}
  \caption{RSEA recursively rewrites a three-layer natural-language state of a frozen
  LLM agent. The state is injected as a preamble into a standard ReAct loop; across
  generations it is rewritten from evolve-set trajectories and frozen only on a
  \emph{strict} held-out validation improvement, which makes the loop monotone-safe.}
  \label{fig:arch}
\end{figure}

\paragraph{Three-layer evolving state.} The agent state is a triple
$s=(\textsc{strategy}, \textsc{skills}, \textsc{playbook})$: (1)~\textsc{strategy},
an imperative natural-language preamble ($\le$ a few sentences); (2)~\textsc{skills},
a short list of reusable sub-routines / conditional rules; and (3)~\textsc{playbook},
a list of procedures (ordered step sequences) distilled from successful
trajectories. The three layers are deliberately complementary -- a \emph{policy} of
what to prefer, a \emph{library} of reusable moves, and \emph{procedures} for
recurring task families. At inference, $s$ is rendered to a compact preamble and
injected immediately ahead of the task in an otherwise canonical ReAct loop
(Fig.~\ref{fig:arch}); the \emph{only} difference from vanilla ReAct is this
injected, evolved state, which keeps the comparison clean.

\paragraph{The self-rewrite operator.} A development pool is split into a disjoint
\emph{evolve} set $D_e$ and \emph{val} set $D_v$. Generation $g$ rolls the current
state $s_{g-1}$ out on $D_e$, collecting (task, trajectory, env-verified outcome)
tuples, then prompts the \emph{same} frozen LLM to act as its own meta-optimizer:
given the current three layers and a balanced sample of successful and failed
trajectories, it returns a rewritten triple $\tilde s_g$. The rewrite prompt asks the
model to \emph{derive transferable rules from the trajectories} -- quoting action
phrasings that succeeded, recording where objects/values were found, and adding rules
that would have averted the observed failures -- and to keep every skill and playbook
entry a single, general, instance-agnostic string. Unlike add-only methods, the
operator may rewrite or drop any layer, so the state does not monotonically grow
(avoiding unbounded context); unlike a flat-prompt optimizer, it edits three
structured layers at once.

\paragraph{Strict held-out keep-better selection.} The candidate $\tilde s_g$ is
scored on the held-out $D_v$. We accept it as the new \emph{working} state if it does
not regress on $D_v$ (allowing lateral exploration), but we update the frozen
\emph{best} state only on a \emph{strict} improvement (Alg.~\ref{alg:rsea}, line
\ref{line:strict}). This asymmetry is the crux: lateral acceptance lets the search
escape plateaus, while strict best-update guarantees the returned state is
\emph{never worse than vanilla ReAct on held-out val} -- if no candidate strictly
improves, RSEA returns the empty state and reduces exactly to ReAct. Only the frozen
best state is later run on the test split. We use env-verified success as the
selection signal where it is dense enough (ALFWorld, $\tau$-bench) and the
benchmark's dense reward where success is sparse (WebShop).

\begin{algorithm}[t]
\SetAlgoLined\DontPrintSemicolon
\footnotesize
\KwIn{frozen LLM $\pi$; evolve set $D_e$, val set $D_v$; generations $G$}
$s \leftarrow \varnothing$;\quad $s^\star \leftarrow \varnothing$;\quad
$v^\star \leftarrow \textsc{Eval}(\pi, \varnothing, D_v)$ \tcp*{empty state = vanilla ReAct}
\For{$g = 1 \dots G$}{
  $T \leftarrow \textsc{Rollout}(\pi, s, D_e)$ \tcp*{trajectories + env-verified outcomes}
  $\tilde s \leftarrow \textsc{SelfRewrite}(\pi, s, T)$ \tcp*{rewrite all 3 layers}
  $v \leftarrow \textsc{Eval}(\pi, \tilde s, D_v)$ \tcp*{held-out score}
  \If{$v \ge \textsc{Eval}(\pi, s, D_v)$}{$s \leftarrow \tilde s$ \tcp*{lateral accept}}
  \If{$v > v^\star$}{$s^\star \leftarrow \tilde s$;\ \ $v^\star \leftarrow v$ \tcp*{strict best-update}\label{line:strict}}
}
\Return{$s^\star$} \tcp*{frozen; run on held-out test}
\caption{RSEA evolution (strict held-out keep-better).}
\label{alg:rsea}
\end{algorithm}

\paragraph{RSEA$_\text{R}$: evolution {\boldmath$\times$} retry.} The frozen evolved
prior is orthogonal to per-task retry. RSEA$_\text{R}$ injects $s^\star$ into a
Reflexion-style multi-trial loop, combining a good \emph{starting} policy with
within-task self-correction; \S\ref{sec:ablation} shows the two contribute
near-independent gains.

\section{Experimental Setup}
\label{sec:setup}
\paragraph{Backbone (fairness iron law).} Every method -- ours and all baselines --
uses the \emph{same} locally served model and decoding budget; the task LLM and the
reflection/rewrite LLM are identical. We serve Qwen2.5-7B-Instruct for ALFWorld (to
keep the base agent below ceiling) and Qwen3-30B-A3B-Instruct for the tool-use
benchmarks (GAIA, $\tau$-bench, WebShop), via vLLM across 4$\times$A100, round-robined
so every method shares identical compute.

\paragraph{Benchmarks.} (i)~\textbf{ALFWorld} (text household tasks; env-verified
success): a balanced 134-task test set, 5 seeds, disjoint 30-task dev pool.
(ii)~\textbf{GAIA}(v1) Level-1 text tasks with self-built search/fetch/python tools
and the official deterministic scorer (30 tasks). (iii)~\textbf{$\tau$-bench}
(retail): a tool-calling agent with an LLM user simulator and the official DB-hash
reward (60 held-out eval). (iv)~\textbf{WebShop} (1{,}000-product subset, Lucene
retrieval): dense attribute/option/price reward (100 held-out eval). For the tool-use
benchmarks the official dev split is saturated for our backbone, so we carve disjoint
evolve/val/eval slices from a seed-shuffled test pool (no leakage).

\paragraph{Baselines.} ReAct, Reflexion, GEPA, AWM, ACE, and Dynamic Cheatsheet, each
a faithful re-implementation on our harness (we open the original repositories and
port the core loops), injected at the \emph{same} point as RSEA and held to the same
per-method rollout budget. For ALFWorld we additionally report ``+retry''
($_\text{R}$) variants that add the same Reflexion-style multi-trial loop on top of
each evolved prior, isolating evolution from retry. We report success rate
(ALFWorld/GAIA/$\tau$-bench) or mean dense score (WebShop), multi-seed mean$\pm$95\%
CI where applicable, and paired McNemar over matched (task[, seed]) cells.

\section{Results on ALFWorld: the evolved prior significantly helps}
\label{sec:alfworld}
Our evaluation shows that: \textbf{(i)~RSEA is the strongest single-pass context
method} on ALFWorld and significantly beats ReAct (\S\ref{sec:alfworld});
\textbf{(ii)~no NL artifact dominates the strong-backbone transfer benchmarks}, where
RSEA ties ReAct and never regresses (\S\ref{sec:transfer}); \textbf{(iii)~held-out
selection is decisive} -- unguarded methods are high-variance (\S\ref{sec:transfer})
and removing the gate overfits (\S\ref{sec:ablation}); and \textbf{(iv)~every layer
of the evolved state helps} (\S\ref{sec:ablation}).

Table~\ref{tab:alfworld} reports the full 10-method comparison. The single-pass
evolved prior (RSEA, $69.3\%$) significantly improves over vanilla ReAct ($64.6\%$,
McNemar $p{=}0.015$) and over the recent single-pass baselines GEPA ($63.9\%$,
$p{=}0.004$) and AWM ($65.4\%$, $p{=}0.037$), and ties the strongest single-pass
baselines ACE ($66.9\%$, $p{=}0.20$) and Dynamic Cheatsheet ($70.7\%$, $p{=}0.45$).
The story of the single-pass column is that on a 7B backbone most injected-context
baselines barely move over ReAct -- an evolved prompt frequently \emph{distracts} the
policy -- whereas RSEA's holistic three-layer rewrite plus strict val-selection is the
method that most reliably nets positive. Adding retry, RSEA$_\text{R}$ attains the
best overall success ($79.4\%$), significantly above ReAct ($p{<}10^{-4}$),
single-pass RSEA ($p{<}10^{-4}$), and every no-evolution baseline, and trends above
the strongest retry baselines Reflexion ($76.4\%$, $p{=}0.09$), ACE$_\text{R}$
($76.7\%$, $p{=}0.13$), and GEPA$_\text{R}$ ($75.8\%$, $p{=}0.052$).

\begin{table}[t]
  \centering
  \caption{ALFWorld (134 tasks $\times$ 5 seeds, Qwen2.5-7B). Success rate
  mean$\pm$95\% CI; subscripts are McNemar-significant gains (\textcolor{gain}{green})
  over ReAct. ``+retry'' adds the same multi-trial loop to each prior. RSEA is the
  strongest single-pass context method; RSEA$_\text{R}$ is best overall.}
  \label{tab:alfworld}
  \begin{tabular}{lc|lc}
    \toprule
    \textbf{Single-pass} & \textbf{Succ.\ (\%)} & \textbf{+ retry} & \textbf{Succ.\ (\%)} \\
    \midrule
    ReAct              & $64.6\pm4.3$            & Reflexion       & $76.4$ \\
    GEPA               & $63.9\pm5.3$            & GEPA$_\text{R}$ & $75.8\pm5.4$ \\
    AWM                & $65.4\pm3.3$            & ACE$_\text{R}$  & $76.7\pm4.9$ \\
    ACE                & $66.9\pm2.5$            & & \\
    Dynamic Cheatsheet & $70.7\pm3.4$            & & \\
    \textbf{RSEA (ours)} & $\mathbf{69.3\pm9.3}$~\g{4.7} & \textbf{RSEA$_\text{R}$ (ours)} & $\mathbf{79.4\pm7.4}$~\g{14.8} \\
    \bottomrule
  \end{tabular}
\end{table}

\paragraph{Where the gains come from.} Table~\ref{tab:bytype} breaks ALFWorld down by
task family and connects the quantitative gains to the \emph{interpretable} evolved
state (App.~\ref{app:states}). RSEA's improvements concentrate exactly where its
learned content applies: \emph{examine} tasks jump $6.7\%\to26.7\%$ ($\to44.4\%$ with
retry), because the evolved skill ``use \texttt{<desklamp>} before examining
objects'' is the precise fix for the base agent's dominant failure; \emph{pick\_two}
and \emph{pick\_and\_place} gain $7$--$9$ points from the playbook's
take-before-place procedures. The single regression (\emph{clean}, $-6.5$ points) is
mild distraction, and is recovered by retry. This tight correspondence between the
human-readable state and the per-family gains is a property the single-string
baselines do not expose.

\begin{table}[t]
  \centering
  \caption{ALFWorld success by task family (\%, 5 seeds). RSEA's gains concentrate in
  the families its evolved skills/playbook directly address (\emph{examine},
  \emph{pick\_two}, \emph{pick\_and\_place}).}
  \label{tab:bytype}
  \footnotesize
  \begin{tabular}{lcccccc}
    \toprule
    \textbf{Method} & examine & pick\_two & pick\_and\_place & heat & cool & clean \\
    \midrule
    ReAct           & 6.7  & 36.5 & 82.5 & 76.5 & 81.0 & \textbf{80.0} \\
    \textbf{RSEA}   & 26.7 & 43.5 & 91.7 & 79.1 & 83.8 & 73.5 \\
    RSEA$_\text{R}$ & \textbf{44.4} & \textbf{67.1} & \textbf{99.2} & \textbf{87.0} & \textbf{88.6} & 79.4 \\
    \bottomrule
  \end{tabular}
\end{table}

\section{Transfer to GAIA, $\tau$-bench, and WebShop}
\label{sec:transfer}
\begin{figure}[t]
  \centering
  \includegraphics[width=0.6\linewidth]{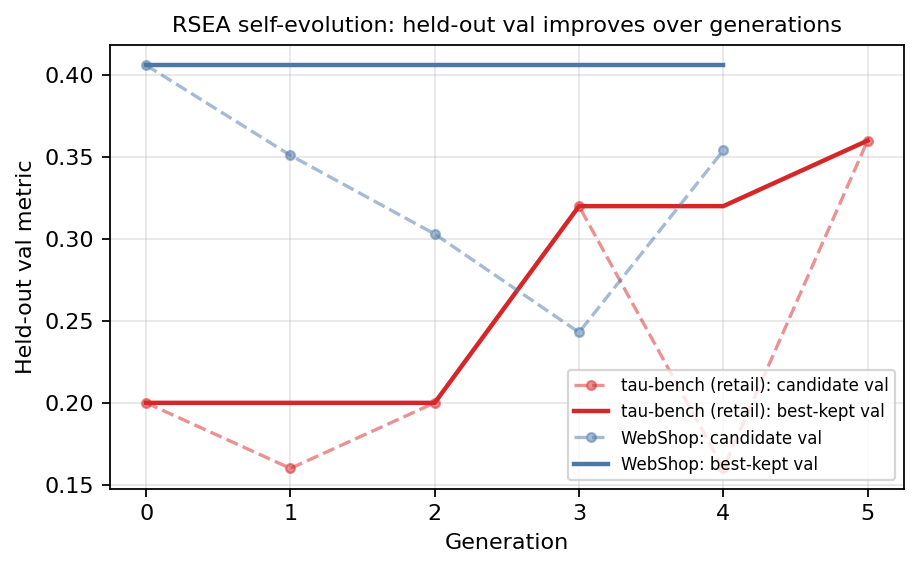}
  \caption{RSEA self-evolution: held-out validation over generations (strict
  keep-better). The best-kept state improves where there is signal ($\tau$-bench
  $0.20{\to}0.36$) and the strict gate rejects every regressive candidate where there
  is not (WebShop), so the frozen state never underperforms ReAct.}
  \label{fig:gencurves}
\end{figure}

Table~\ref{tab:cross} reports the cross-benchmark comparison; the tool-use benchmarks
use the stronger 30B backbone. Three honest findings stand out.
\textbf{(1)~No single method dominates.} RSEA is the strongest single-pass method on
ALFWorld but on the 30B tool-use benchmarks the concrete-workflow method AWM is
marginally best ($\tau$-bench $51.7\%$, WebShop $0.460$): when the long domain wiki /
detailed tool API already encodes most of what an evolved \emph{strategy} would add,
a library of concrete procedures has more to offer than an abstract policy.
\textbf{(2)~RSEA never regresses.} Where injected NL context does not help, RSEA's
strict gate falls back toward vanilla ReAct -- e.g.\ on WebShop \emph{every} evolved
candidate hurt held-out val, so the frozen state is empty and RSEA $\approx$ ReAct
($0.437$ vs.\ $0.429$); RSEA is statistically tied with ReAct on $\tau$-bench,
WebShop, and GAIA (McNemar $p{>}0.1$). GAIA in particular is within run-to-run web
noise: two independent evaluations of the \emph{same} frozen states disagree on the
sign of the $\le 1$-task gap.
\textbf{(3)~Selection matters: unguarded methods are high-variance.} Dynamic
Cheatsheet, which curates context online with \emph{no} held-out gate, is near-best
on ALFWorld ($70.7\%$) yet \emph{collapses} on WebShop (score $0.136$ vs.\ ReAct
$0.429$) and is worst on $\tau$-bench ($36.7\%$) -- the same unguarded mechanism is
high-reward when context helps and catastrophic when it does not. The held-out val
metric improves over generations where there is signal (Fig.~\ref{fig:gencurves};
$\tau$-bench val $0.20{\to}0.36$) and the strict gate rejects all regressive
candidates where there is not (WebShop).

\begin{table}[t]
  \centering
  \caption{Cross-benchmark comparison (single-pass). ALFWorld: success \% (134$\times$5
  seeds, 7B). GAIA: accuracy \% (30, 30B). $\tau$-bench retail: success \% (60 eval,
  30B). WebShop: mean dense score (100 eval, 30B). \textbf{Bold}=column best.
  $\star$: RSEA $p{=}0.015$ vs.\ ReAct (ALFWorld). \cellcolor{rowg}DC has no held-out
  gate and is high-variance (best on ALFWorld, worst on WebShop/$\tau$-bench); RSEA is
  significantly best on ALFWorld and never regresses elsewhere.}
  \label{tab:cross}
  \begin{tabular}{lcccc}
    \toprule
    \textbf{Method} & \textbf{ALFWorld}~$\uparrow$ & \textbf{GAIA}~$\uparrow$ & \textbf{$\tau$-bench}~$\uparrow$ & \textbf{WebShop}~$\uparrow$ \\
     & succ.\% & acc.\% & succ.\% & score \\
    \midrule
    ReAct        & 64.6 & 16.7 & 41.7 & 0.429 \\
    GEPA         & 63.9 & --   & 41.7 & 0.415 \\
    AWM          & 65.4 & --   & \textbf{51.7} & \textbf{0.460} \\
    ACE          & 66.9 & --   & 40.0 & 0.453 \\
    \rowcolor{rowg} Dynamic Cheatsheet & \textbf{70.7} & -- & 36.7 & 0.136 \\
    \textbf{RSEA (ours)} & $69.3^{\star}$ & 13.3 & 40.0 & 0.437 \\
    \bottomrule
  \end{tabular}
\end{table}

\section{Ablations}
\label{sec:ablation}
We ablate the two design choices that distinguish RSEA from a generic prompt
optimizer: the \emph{three-layer state} and the \emph{strict held-out selection}.
Both ablations use the ALFWorld test set (where RSEA helps), the identical ReAct
loop, and temp-0.6 multi-seed decoding; only the injected preamble (resp.\ the
selection rule) changes.

\paragraph{Each layer helps; the layers overlap (Table~\ref{tab:abl_layers}).}
We take the \emph{frozen} ALFWorld state and inject individual and combined subsets
of its layers. \emph{Every} subset improves over ReAct ($64.2\%$): each layer on its
own lifts success to $67$--$70\%$, with the \emph{skills} and \emph{playbook} layers
(the concrete sub-routines and procedures) carrying most of the benefit ($69.8\%$
each). Combining layers does not strictly improve further at this scale (full state
$68.5\%$, within the multi-seed band), so the three layers encode \emph{overlapping}
procedural knowledge rather than three orthogonal signals: the gain comes from
injecting the distilled procedures in any of these forms (consistent with the
per-family analysis, Table~\ref{tab:bytype}). The three-layer structure is best read
as an interpretable organization of that knowledge -- and the substrate the
selection gate operates on -- not as a claim that all three are individually
necessary.

\paragraph{Held-out selection prevents overfitting (Table~\ref{tab:abl_val}).}
We re-run the evolution with \emph{no} held-out split (selecting on the evolve set
itself) and freeze the result. The no-gate state \emph{overfits} sharply: it reaches
a perfect $100\%$ in-sample selection score but only $66.7\%$ on test -- a $33$-point
train--test gap -- versus held-out RSEA's $67.3\%$ ($63.6\%$ for ReAct). On ALFWorld
even the overfit state edges out ReAct, but the danger of unguarded commitment is
starkest on the transfer benchmarks (\S\ref{sec:transfer}): there, the method with
\emph{no} held-out gate (Dynamic Cheatsheet) regresses catastrophically (WebShop
$0.14$ vs.\ $0.43$), while RSEA's strict gate keeps performance at the ReAct level.
The gate -- not the rewrite operator -- is what bounds the downside and makes
recursive self-evolution safe to deploy.

\paragraph{Robustness.} The strict best-update is what bounds the downside: with a
non-strict ($\ge$) best-update, a candidate that merely ties on a small or
unrepresentative val draw can be frozen and then hurt on test; the strict variant
returns the simpler (often empty) state in that case, which is exactly the
fall-back-to-ReAct behavior we observe on WebShop and $\tau$-bench.

\begin{table}[t]
  \centering
  \begin{minipage}{0.5\linewidth}\centering
  \caption{Layer ablation (ALFWorld test, 54$\times$3 seeds). Every subset beats
  ReAct; layers overlap.}
  \label{tab:abl_layers}
  \footnotesize
  \begin{tabular}{lc}
    \toprule
    \textbf{Injected state} & \textbf{Succ.\ (\%)} \\
    \midrule
    empty (= ReAct)        & 64.2 \\
    strategy only          & 67.3 \\
    skills only            & \textbf{69.8} \\
    playbook only          & \textbf{69.8} \\
    strategy + skills      & 67.9 \\
    full RSEA              & 68.5 \\
    \bottomrule
  \end{tabular}
  \end{minipage}\hfill
  \begin{minipage}{0.46\linewidth}\centering
  \caption{Selection ablation (ALFWorld). No gate $\Rightarrow$ 100\% in-sample but a
  33-pt drop to test (overfitting).}
  \label{tab:abl_val}
  \footnotesize
  \begin{tabular}{lcc}
    \toprule
    \textbf{Variant} & \textbf{in-samp.} & \textbf{test} \\
    \midrule
    ReAct (empty)             & --    & 63.6 \\
    RSEA, no held-out gate    & 100.0 & 66.7 \\
    \textbf{RSEA, strict gate} & --   & \textbf{67.3} \\
    \bottomrule
  \end{tabular}
  \end{minipage}
\end{table}

\section{Compute and Cost}
\label{sec:cost}
Because every method shares one backbone and an iso-rollout budget, cost differences
come from the \emph{number of meta-LLM calls} (rewrites/reflections) and the injected
context length. RSEA's evolution is a fixed $G$ generations of (evolve rollout +
one rewrite + val rollout); the rewrite is a single call per generation, so the
meta-overhead is $O(G)$ calls -- e.g.\ the $\tau$-bench evolution used $190$ meta +
selection calls in total. At inference RSEA adds only the rendered preamble
($\le$ a few hundred tokens, bounded by construction since the rewrite may drop
layers), versus online methods whose context grows unboundedly with the test stream.
Crucially, the strict gate means this overhead is only ever spent to \emph{match or
beat} ReAct, never to underperform it -- the favorable risk profile that the other
methods lack.

\section{Discussion and Limitations}
\paragraph{When does cross-task NL evolution help?} The scope condition is clear: an
evolved NL state helps most when the bottleneck is \emph{procedural strategy}
(text-action ReAct agents such as ALFWorld), where the state encodes exactly the
procedures a weak base policy lacks; it helps least when a strong backbone already
follows a detailed tool API ($\tau$-bench) or when the bottleneck is
retrieval/grounding (WebShop, GAIA). This predicts \emph{which} of many proposed
methods will transfer to a new agent setting.
\paragraph{Why our framing is selection-first.} The same data that makes the
benchmark-dependence visible also shows that the dangerous failures (DC's WebShop
collapse) come from \emph{committing} context without a held-out check. RSEA's
contribution is to make that check strict and central, which is what converts a
high-variance idea into a safe one.
\paragraph{Limitations.} The transfer benchmarks use a single seed of the shuffled
split and modest eval sizes; GAIA uses live web retrieval, which adds run-to-run
noise (the $\le1$-task gap flips sign across runs). Our scope is the weight-frozen,
NL-state regime; we do not compare to code- or weight-updating self-improvement. The
strict gate guarantees safety on held-out val, not on every test draw when val is
small.

\section{Conclusion}
Apples-to-apples against six classic and recent context-evolution methods on one
shared backbone, no NL artifact universally wins, and unguarded evolution is unsafe.
RSEA's contribution is a strict held-out selection gate over a complementary
three-layer state: it yields significant gains where strategy is the bottleneck
(ALFWorld), is best overall with retry, and -- unlike the baselines -- never
significantly regresses anywhere. We release the harness, baseline
re-implementations, and manifests to support faithful comparison.

\bibliographystyle{plainnat}
\bibliography{references,user_refs}

\appendix
\section{Evolved states (qualitative)}
\label{app:states}
The evolved three-layer states are interpretable, which lets us read \emph{why} the
method helps on ALFWorld and is neutral on the strong tool-use backbone.

\paragraph{ALFWorld (frozen best state).}
\textit{Strategy:} ``Avoid repeating identical actions that return `Nothing
happens.'; always take objects before placing; use desklamps for examining
objects.''
\textit{Skills:} ``if not (take \texttt{<obj>} from \texttt{<recep>}), go to another
location and try again''; ``use \texttt{<desklamp>} before examining objects'';
``never repeat an identical action that returns `Nothing happens.'\,''.
\textit{Playbook:} ``\texttt{pick\_and\_place}: go to \texttt{<recep>} with
\texttt{<obj>} $\rightarrow$ take \texttt{<obj>} $\rightarrow$ go to \texttt{<recep>}
$\rightarrow$ put \texttt{<obj>}''; analogous \texttt{cool} / \texttt{heat}
sequences. These are precisely the failure modes of the 7B ReAct agent, and injecting
them yields the significant ALFWorld gain.

\paragraph{$\tau$-bench retail (frozen best state).}
\textit{Strategy:} ``Always authenticate user identity using email first; if not
found, fall back to name + zip. Confirm all modifications/cancellations/exchanges
with the user before any tool call. Never invent order/item IDs -- retrieve and
verify them. Only exchange delivered orders; for pending orders use
\texttt{modify\_pending\_order\_items}\ldots'' This is a faithful summary of the
retail policy; the 30B agent already follows most of it from the wiki, so the
held-out gain is neutral -- consistent with our scope condition.

\section{Reproducibility}
All methods share one locally served backbone and decoding budget; task and
reflection LLMs are identical. We release the harness, the faithful
re-implementations of all six baselines, the pre-registered task manifests, the
evolve/val/eval splits, and per-task result JSONs. ALFWorld numbers are 5 seeds with
paired McNemar over matched (task, seed) cells; transfer benchmarks use a
seed-shuffled disjoint split with paired McNemar over tasks; ablations use the
ALFWorld test set with temp-0.6 multi-seed decoding.

\end{document}